\documentclass{bmvc2k}

\usepackage{times}
\usepackage{multirow}
\usepackage{epsfig}
\usepackage{graphicx}
\usepackage{amsmath}
\usepackage{amssymb}
\usepackage{array}
\usepackage{soul}
\usepackage{setspace}
\usepackage{float}
\usepackage{algorithm2e}
\usepackage{capt-of}
\usepackage{subfigure}

\SetKwProg{Fn}{Function}{}{}
\DeclareMathOperator*{\argmax}{arg\,max} 

\title{Local Visual Microphones: Improved Sound Extraction from Silent Video}

\addauthor{Mohammad Amin Shabani}{shabani_mohammadamin@dena.sharif.ir}{1}
\addauthor{Laleh Samadfam}{samadfam@ce.sharif.ir}{1}
\addauthor{Mohammad Amin Sadeghi}{asadeghi@ut.ac.ir}{2}

\addinstitution{
	Sharif University of Technology\\
	Azadi Avenue\\
	Tehran, Iran
}
\addinstitution{
	University of Tehran\\
	Tehran Province, 16 Azar St, Enghelab Square
	Tehran, Iran
}
\runninghead{Shabani, Samadfam, Sadeghi}{Local Visual Microphones}

\def\etal{\emph{et al}\bmvaOneDot}

\begin{document}
	
	\maketitle
	
	\begin{abstract}
		Sound waves cause small vibrations in nearby objects. A few techniques exist in the literature that can extract sound from video. In this paper we study local vibration patterns at different image locations. We show that different locations in the image vibrate differently. We carefully aggregate local vibrations and produce a sound quality that improves state-of-the-art. We show that local vibrations could have a time delay because sound waves take time to travel through the air. We use this phenomenon to estimate sound direction. We also present a novel algorithm that speeds up sound extraction by two to three orders of magnitude and reaches real-time performance in a 20KHz video.
		
	\end{abstract}
	
	%-------------------------------------------------------------------------
	%%%%%%%%% BODY TEXT
	\begin{figure}[h]
		\begin{center}
			{\includegraphics[scale=.18]{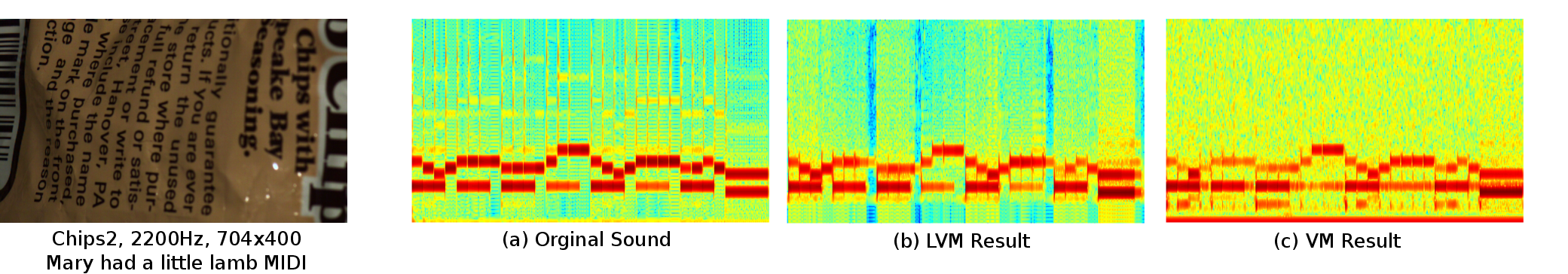}}    
		\end{center}
		\vspace*{-5mm}
		\caption{Left: Input high-speed video, a: Spectrogram of original sound. b: Spectrogram of our recovered sound. c: Spectrogram of sound recovery by~\cite{Davis2014VisualMic}.}
		\label{fig:splash}
	\end{figure}
	\vspace*{-5mm}
	\section{Introduction}
	
	Sound is a wave of pressure and displacement through air or a different medium. When sound waves hit an object, they cause tiny vibrations on the objects. These vibrations are sometimes detectable. Sound extraction techniques~\cite{wang2014audio, Davis2014VisualMic,zalevsky2009simultaneous} work by measuring tiny displacements over time in a high-speed video. Displacements as low as one hundredth of a pixel can be detected in good conditions~\cite{Guizar-Sicairos:08,zhang2016high,foroosh2002extension}.
	
	Sound extraction techniques prior to this work either measure global image displacement~\cite{zhang2016high, wang2014audio} or measure local displacements and compute a weighted average according to the richness of texture~\cite{Davis2014VisualMic}. A texture-rich area in the image does not necessarily detect a stronger sound signal because multiple other factors affect signal quality.
	
	Different areas in a scene react differently to incoming sound. In fact, the way an object reacts to sound strongly depends on a number of factors: material, object geometry, sound frequency and distance. Also, the way vibrations are visible to camera depends on a few other factors: texture, direction with respect to image plane, and edge direction. We study these effects in Section~\ref{sec:localsound} .
	
	Due to the above factors different areas of the image will have vibrations with different amplitudes, phases and frequency characteristics. As a result, simply averaging local vibrations will distort sound quality. One of the contributions of this paper is to examine vibrations in local areas and generate global sound by carefully aggregating vibrations in local areas. As a result, We improve the state-of-the-art of sound extraction quality. As a byproduct of our local vibration analysis, we can extract further information from silent video including the direction of the incoming sound.
	%*****************************
	\begin{figure}[t]
		\begin{center}
			\scalebox{.3}{\includegraphics{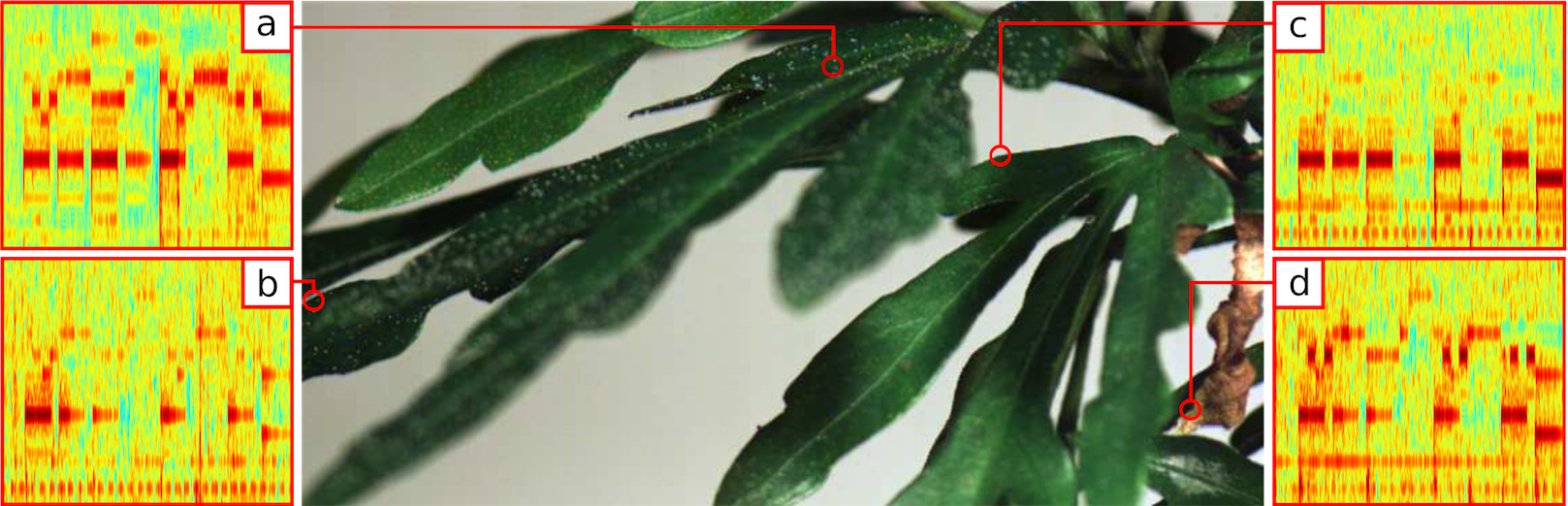}}    
			\caption{Different locations in the image vibrate differently and capture different aspects of the sound. Each one of the four spectrograms is from a different location. Please note that these four locations have different sound qualities and frequency characteristics.}
		\end{center}
		\label{fig:specOfPatches}
	\end{figure}
	
	%*****************************
	Our experiments show that a there is a small subset of all pixels that carry most of the sound information. Most pixels carry little sound information. Another benefit of local sound extraction is that we can aggressively cut down computation cost by limiting computation to only the most useful pixels (processing high speed video is computationally intensive).
	
	We developed a very fast algorithm to extract sound. Our algorithm performs less than 20 arithmetic operations per every pixel that is processed. As a result, our vibration detection algorithm has a throughput of about 1 Gigapixels per second on a commodity computer. As a result, our algorithm runs in real-time on a 20KHz video.
	
	\section{Prior Work}\label{sec:prior}
	Motion extraction from a video has a wide range of applications. In the 1980s, the first applications of motion extraction were in biomedical imaging~\cite{althof1997rapid, alves2015computer} and video compression~\cite{sullivan1991motion}. Today, motion extraction has more diverse applications. In health, Zeev \etal~\cite{zalevsky2009simultaneous} extracted heartbeats by detecting tiny movements using an optical interferometer. Balakrishnan \etal~\cite{6619284}, attempted to detect heart rate by tracking a number of key-points on a patient's head.
	
	In civil engineering, Young-Jin \etal~\cite{cha2015motion} used motion magnification to detect structural damages in urban structures. In physics, Xue \etal~\cite{Xue2014} used motion detection to estimate the speed of moving hot air or similar transparent fluids against a textured background. In material engineering, Davis \etal~\cite{7299171} used sub-pixel motion extraction to estimate certain material properties from a video~\cite{7299171}. In 3D video processing, motion detection is used to extract depth map from binocular images~\cite{ideses2007real, chang2007depth}. 
	
	For sound detection, Zeev \etal~\cite{zalevsky2009simultaneous} 
	extracted speech signals by measuring the vibration of people's neck in a video. 
	Davis \etal~\cite{Davis2014VisualMic} used sub-pixel motion extraction to recover sound from silent video. They extract vibration separately in a number of scales and angles. Then they align signals temporally (to avoid destructive interference) and finally they take a weighted average among all orientations and scales.
	
	%%%%%%%%%%%%%%%%%%%%%%%%%%%%%%%%%%%%%%%%%%%%%%%%%%%%%%%%%%%%
	Before we go over further details of prior work, we need to establish relevant notations and assumptions. We assume we are given an input video $F$, that refers to a sequence of $l$ frames $F_1 \ldots F_l$. For simplicity, we assume video frames are 1-D gray-scale vectors as opposed to 2-D color matrices. We assume each frame $F_t$ ($1 \leq t \leq l$) is a vector of length $m$. Here, $F_t(x)$ refers to the intensity of the $x$'th pixel in the $t$'th frame~($x,t\in\mathbb{N}$). Figure 4 illustrates our notation in more details. 
	
	In order to extract motion, we need to compute displacement over time. We measure displacement with respect to a reference frame $F_{r}$. We define displacement in $F_t$ as $d_r(t)$ (where $r$ identifies reference frame) or simply $d(t)$.
	A simple way to measure $d(t)$ is to use cross-correlation:
	\begin{equation}
	d_r(t) = \argmax_{\Delta x} \sum_{x} F_r(x)F_t(x-\Delta x) 
	\label{equation:cc}
	\end{equation}    
	Basic cross-correlation gives a precision of one pixel. In applications where displacement is significantly larger than one pixel, variants of cross-correlation~\cite{anandan1989computational,212789,kalivas1991region} or feature point tracking~\cite{Liu:2005:MM:1073204.1073223, zalevsky2009simultaneous} often produce satisfactory precision. However, in cases where displacement is less than one pixel, sub-pixel precision algorithms are needed.
	
	A number of approaches to compute sub-pixel displacement have been proposed. These approaches generally interpolate cross-correlation to sub-pixel precision and calculate $\Delta x$ in sub-pixel precision~\cite{tian1986algorithms, Guizar-Sicairos:08, zhang2016high}.
	
	Guizar-Sicairos \etal~\cite{Guizar-Sicairos:08} first compute displacement with a precision of one pixel using cross-correlation. Then, they further refine displacement using interpolation. After finding motion in pixel precision, Zhang \etal~\cite{zhang2016high} proposed two different methods to refine $d(t)$. They use Taylor approximation to interpolate cross-correlation to sub-pixel precision. 
	
	Another approach is to apply a bandpass filter on a sequence of pixel intensities and extract frequency bands of interest~\cite{wu2012eulerian, wadhwa2013phase, Elgharib2015VideoMI, foroosh2002extension}. Wadhwa \etal~\cite{wadhwa2013phase} first decompose images into different spatial scales and orientations. Then, they use complex-steerable-pyramids to separate the amplitude of local wavelets. Finally, they apply a temporal filter on phases at each location, orientation, and scale.
	%%%%%%%%%%%%%%%%%%%%%%%%%%%%%%%%%%%%%%%%%%%%%%%%%%%%%%%%%%%%
	%%%%%%%%%%%%%%%%%%%%%%%%%%%%%%%%%%%%%%%%%%%%%%%%%%%%%%%%%%%%
	%%%%%%%%%%%%%%%%%%%%%%%%%%%%%%%%%%%%%%%%%%%%%%%%%%%%%%%%%%%%    
	\section{Motion Extraction Algorithm}\label{sec:algorithm}
	%*****************************
	\begin{figure}[h]
		\begin{center}
			{\includegraphics[width=1.01\textwidth]{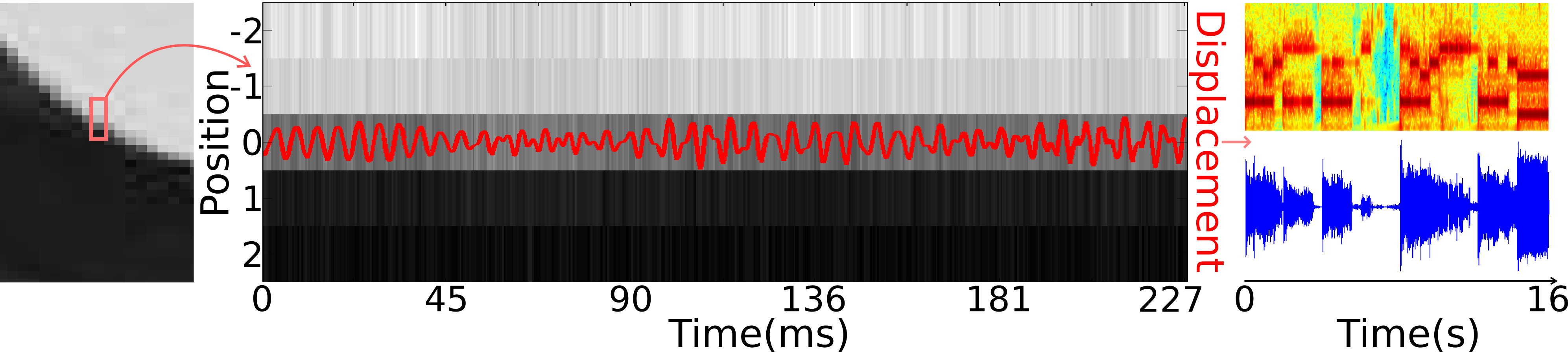}}    
		\end{center}
		\vspace*{-5mm}
		\caption{An illustration of sound extraction process. Left: a $5\times 1$ pixels location in the image. Middle: the value of these five pixels over a fraction of a second. In this figure pixel intensity variations are exaggerated five times and displacement (red curve) is exaggerated 50 times. Right: Extracted sound from video. Note that vibration amplitude could be as small as one-hundredth of a pixel. We capture this tiny displacement at several locations to extract sound.}
		
	\end{figure}
	%*****************************   
	
	Given $F_r$, $F_t$, and $\Delta x$, we define operator $K$ as follows:
	
	\begin{equation}
	K(F_r, F_t, \Delta x ) = \sum_{x} F_r(x)F_t(x-\Delta x) ~~~~~~~ \forall \Delta x \in \mathbb{Z}.
	\label{equation:kernel}
	\end{equation}
	
	Operator $K$ displaces $F_t$ by $\Delta x$ and then computes its similarity with $F_r$. As noted in Equation~\ref{equation:cc}, to find the displacement between $F_t$ and $F_r$, we find a displacement $\Delta x$ that maximizes $K(F_r, F_t, \Delta x )$.
	\begin{equation}
	\argmax_{\Delta x} K(F_r, F_t, \Delta x );
	\label{equation:argmax}
	\end{equation}    	
	For $\Delta x \in \mathbb{Z}$ calculations of Equations~\ref{equation:kernel} and~\ref{equation:argmax} are straightforward.
	To extend $K(F_r, F_t, \Delta x)$ for real values of $\Delta x$ we can interpolate this function (other alternatives to interpolation were discussed in Section~\ref{sec:prior}). To compute displacement with sub-pixel precision, we first estimate the integer part of displacement using Equation~\ref{equation:argmax}, and then estimate sub-pixel fraction using interpolation.
	
	We evaluate $K(F_r,F_t,\Delta X)$ for $\Delta x \in \{-1, 0, 1\}$ and define $\hat{K}$ for real values of $-1 \leq \Delta x \leq  1 $ as a quadratic function:
	\begin{equation}
	\hat{K}(F_r, F_t, \Delta x ) ~=~ a {\Delta x}^2 ~+~ b \Delta x ~+~ c ~~~~~~~~~~~ \forall \Delta x \in \mathbb{R}, -1 \leq \Delta x \leq  1   	
	\label{equation:kernel2}
	\end{equation}
	
\begin{figure}[htp]
		\begin{minipage}{0.45\textwidth}
			\centering
			\begin{algorithm}[H]
				
				\Fn{$\argmax_{\Delta x}K(F_r, F_i, \Delta x)$}{
					\SetAlgoLined
					$f(0) = K(F_r, F_i, 0);$\\
					$f(1) = K(F_r, F_i, 1);$\\
					$f(-1) = K(F_r, F_i, -1);$\\
					$d = \dfrac{f(-1) - f(1)}{2f(1) + 2f(-1) - 4f(0)};$\\
					\KwRet d;\\
				}			
			\end{algorithm}
		\end{minipage}
		\hfill
		\begin{minipage}{0.45\textwidth}
			\centering
			\includegraphics{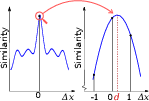}
			\label{fig:interpolation}
		\end{minipage}
		\vspace*{+2mm}
		\caption{Left: Simple pseudo-code to estimate displacement according to similarity function $K(F_r, F_i, \Delta x)$. Right: An illustration of cross-correlation around $\Delta x=0$ and how we estimate displacement using a quadratic interpolation. }
	\end{figure}

	To estimate sub-pixel displacement, we estimate $a,b,c$ and choose a $\Delta x$ that maximizes $\hat{K}$. Figure~\ref{fig:interpolation} illustrates this process in more details.
	\noindent
	\noindent	
	\subsection{Further Considerations in Motion Extraction}
	
	\textbf{Measure of Similarity:} For image-registration and template-matching, often cross-correlation is normalized to remove the effects of magnitude of intensity, Equation~\ref{equation:Nk}.
	
	\begin{equation}
	\bar{K}(F_r, F_t, \Delta x ) = {\mathlarger{\mathlarger{\sum}}_{x}\frac{ F_r(x)F_t(x-\Delta x)}{\sigma_r\sigma_t}}
	\label{equation:Nk}
	\end{equation}
	
	\textbf{Quartic Interpolation:} We can use a quartic interpolation as opposed to quadratic interpolation. For quartic interpolation we use five points instead of 3 points. In Section~\ref{sec:Experimental}, we compare the two techniques.
	
	\textbf{Tracking:} 	We discussed computation of displacement for the case $ |\Delta x| < 1$. In cases where $ |\Delta x| > 1$, we need to first approximate the integer part of displacement. In a high-speed camera, displacement between two pixels is very small. In our case it is less than a hundredth of a pixel most of the time. Therefore, we track displacement and reuse its integer part. In fact the integer part of displacement can move by at most one pixel between two consecutive frames.
	
	\textbf{Two dimensional:}  In Section~\ref{sec:algorithm}, for simplicity, we supposed that frames are one-dimensional signals. However, We can use a two-dimensional version of $\bar{K}$ and estimate both vertical and horizontal displacements at once.
	
	\textbf{Global vs. Local:} As $\Delta x$ varies in different parts of the image, instead of calculating a global displacement signal, we partition the image into several blocks and compute displacement in each block separately. We discuss this in more details in Section~\ref{sec:localsound}. 
	%%%%%%%%%%%%%%%%%%%%%%%%%%%%%%%%%%%%%%%%%%%%%%%%%%%%%%%%%%%%
	%%%%%%%%%%%%%%%%%%%%%%%%%%%%%%%%%%%%%%%%%%%%%%%%%%%%%%%%%%%%
	%%%%%%%%%%%%%%%%%%%%%%%%%%%%%%%%%%%%%%%%%%%%%%%%%%%%%%%%%%%%
	\section{Local Sound Extraction}\label{sec:localsound}
	
	When a sound is played in a scene, different areas in the scene react differently to the incoming sound, Figure~\ref{fig:specOfPatches}. In fact, a number of factors affect how a certain point on an object vibrates given an incoming sound: 
	\begin{itemize}
		\item \textbf{Material:} Some materials react more strongly to incoming sound than other materials. Generally, in higher sound frequencies, thinner and lighter objects vibrate more strongly than thicker objects~\cite{Davis2014VisualMic}.
		\item \textbf{Geometry:} In one-dimensional objects (such as a rope or a rod) anti-nodes vibrate more than nodes. Vibration response from two-dimensional objects is more complex.
		\item \textbf{Frequency:} Objects respond to different frequencies differently. Resonance frequencies amplify vibration while other frequencies lead to phase shift.
		\item \textbf{Distance:} In a 20KHz video, sound waves move about 17 millimeters between every two frames. Therefore, objects that are farther to the sound source will experience some delay. Furthermore, a point that is twice farther to the sound source receives one-fourth of power and half of vibration amplitude.
	\end{itemize}
	
	All of the above factors cause some parts of the image to vibrate more or vibrate less. Further, in those parts that really vibrate, their vibration could be invisible due to a few other factors:    
	\begin{itemize}
		\item \textbf{Texture:} Vibrations in texture-rich areas are more visible than in texture-less areas. As a result, in low-texture areas vibrations could be present but they are difficult to detect.
		\item \textbf{Image plane:} Vibrations along image plane are more visible than off-plane vibrations.
		\item \textbf{Edge direction:} Even though edges are strong textures, vibrations that are perpendicular to an edge are more visible than vibrations along an edge.
	\end{itemize}
	
	%%%%%%%%%%%%%%%%%%%%%%%%%%%%%%%%%%%%%%%%%%%%%%%%%%%%%%%%%%%%
	\subsection{Aggregating Local Vibrations to Improve Quality}\label{sec:accuracy}
	
	In practice, we break down the image to small blocks and then extract sound signal in each block separately. Then, we score each block by the quality of its sound signal. Most blocks have low-quality sound signals. We ignore most blocks and only keep the highest scoring ones. We use \textit{voice activity likelihood}~\cite{Brookes1997}, an available software library to assign score to sound signals. We tried $8 \times 8$, $16 \times 16$ and $32 \times 32$ blocks where $8 \times 8$ produced superior results.
	
	Due to the factors mentioned in the previous section, sound extracted from different blocks, have different amplitudes, phase-shifts and frequency characteristics (Figure~\ref{fig:areas}). Due to strong phase-shifts between blocks, simple averaging of sound signals couse destructive interference and cancel out sound signal in some frequencies.
	
	To avoid destructive interference, we combine local vibrations in a careful way. At any given frequency, each block has a different phase-shift. At each frequency, we ignore phase-shifts from all blocks and then add up the remaining signals. We perform this operation for each frequency separately and then produce an aggregate sound signal by combining all frequencies. This technique led to a state-of-the-art sound quality.
	
	Finally, we use a noise reduction technique from~\cite{loizou2005speech}. Our baseline~\cite{Davis2014VisualMic} reports all experimental results after the same noise reduction process. Improvements from speech enhancement is small and our technique significantly outperforms the previous state-of-the-art both before and after noise reduction.
	
	\begin{table}[t]
		\begin{minipage}[b]{0.45\linewidth}
			\centering
			\begin{tabular}{|  >{\centering} m {2.7cm} | c |  }
				\hline
				Algorithm & Complexity \\ \hline
				Cross Correlation & $O(n\sqrt{n})$  \\ \hline
				Up-sampled Cross Correlation & $O(nl \sqrt{nl})$ \\ \hline
				DFT registration~\cite{Guizar-Sicairos:08} & $O(nl \log{nl})$ \\ \hline
				Ours & \textbf{$O(n)$} \\ 
				\hline
			\end{tabular}
			\vspace*{+3mm}
			\caption{Computational complexity for a number of motion extraction algorithms. Our algorithm has linear complexity with respect to the number of pixels that are processed.}
			\label{tab:algorders}
		\end{minipage}\hspace*{+4mm}
		\begin{minipage}[b]{0.45\linewidth}
			\centering
			\includegraphics[scale=0.22]{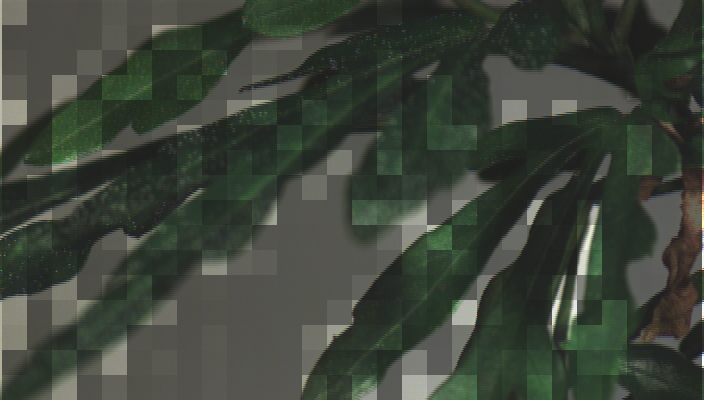}
			
			\captionof{figure}{Sound quality in several image blocks. Brighter patches are estimated to have higher sound quality. Please note that edges capture vibration more strongly.}
			\label{fig:areas}
		\end{minipage}
	\end{table}
	
	%%%%%%%%%%%%%%%%%%%%%%%%%%%%%%%%%%%%%%%%%%%%%%%%%%%%%%%%%%%%
	%%%%%%%%%%%%%%%%%%%%%%%%%%%%%%%%%%%%%%%%%%%%%%%%%%%%%%%%%%%%
	%%%%%%%%%%%%%%%%%%%%%%%%%%%%%%%%%%%%%%%%%%%%%%%%%%%%%%%%%%%%
	
	\section{Real-time performance}\label{sec:time}
	
	Not only does our algorithm beat state-of-the-art sound quality, it is also orders of magnitude faster than previous algorithms. Our algorithm runs in real time and we can cite three reasons for this speed-up.
	
	\begin{itemize}
		\item \textbf{Fast registration algorithms:} Given an $m \times m$ image with $n = m \times m$ pixels, all previous algorithms have super-linear complexities. However, since we track the integer part of displacement and compute similarity function $K$ only three times for each block, our algorithm has a complexity of $O(n)$, Table~\ref{tab:algorders}. In fact our algorithm uses less than 20 floating-point operations per pixel. As a result, our computation complexity is linear and memory bound.	
		
		\item \textbf{Pruning low-quality locations:}  Our experiments show that most of sound signal is concentrated in a few small neighborhoods (Figure~\ref{fig:highschores}). We use techniques to perform computation on only local blocks that are expected to produce high quality sound.
				
		\item \textbf{Hardware optimization:} We use multi-threading, CPU intrinsics and careful cache handling techniques to speed up computation. Unlike complex algorithms like DFT we don't need floating-point arithmetic. Our core computation runs with 16-bit fixed point operations. As a result, we can perform 16 arithmetic operations in each AVX instruction. 
		
	\end{itemize}
	%%%%%%%%%%%%%%%%%%%%%%%%%%%%%%%%%%%%%%%%%%%%%%%%%%%%%%%%%%%%
	%%%%%%%%%%%%%%%%%%%%%%%%%%%%%%%%%%%%%%%%%%%%%%%%%%%%%%%%%%%%
	%%%%%%%%%%%%%%%%%%%%%%%%%%%%%%%%%%%%%%%%%%%%%%%%%%%%%%%%%%%%
	
	\subsection{Sound Extraction with a Few Pixels}\label{sec:mask}
	
	We can prune low-quality pixels in the image more aggressively. We noticed that the intensity of a single pixel over time produces a comprehensible sound. Moreover, taking a weighted average over the intensities of a number of top-rated pixels, gives a moderate sound quality. In Section~\ref{sec:Experimental}, we evaluate the performance of this technique. We assign weights to pixels using voice activation likelihood~\cite{Brookes1997}. Figure~\ref{fig:highschores} shows an image with a number of top-ranking pixels. We compute the mask at first by using a small segment of the video and use it for the rest of the video. 
	
	\section{Sound Directions} \label{sec:orientation}
	
\begin{figure}[t!]
		\begin{minipage}[t]{0.45\textwidth}
			\centering
			
			\scalebox{.43}{\includegraphics{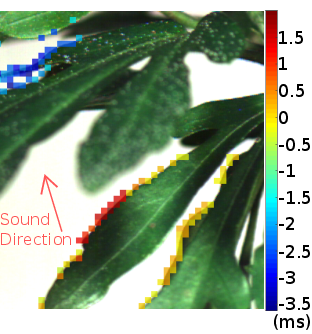}}    
			\caption{Some blocks receive incoming sound slightly earlier than some others. This figure identifies time-shifts for different blocks. We estimate sound direction according to time-shifts.}
			
			\label{fig:direction}
		\end{minipage}
		\hfill
		\begin{minipage}[t]{0.45\textwidth}
			\centering
			\scalebox{.41}{\includegraphics{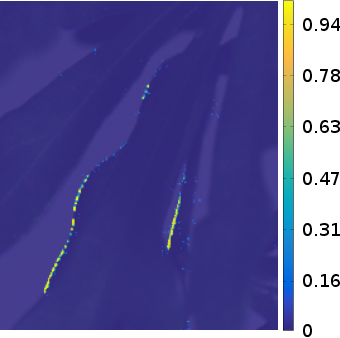}} 	
			\caption{Not all pixels have equal sound quality. This figure visualizes sound quality at various pixels.}
			\label{fig:highschores}
		\end{minipage}
	\end{figure}

	Sound waves travel at a speed of about 340 m/s in air. In a 20KHz video, sound waves travel about 17 millimeters between every two frames. These shifts help us estimate a few factors including sound orientation. To do so, we first extract sound from all blocks and choose high quality ones. Then, we specify one block as $t_0$ and compute time shift for other blocks using cross-correlation. We finally regress a direction vector that shows sound direction. We applied this technique on a few different time intervals and frequencies and they agreed. Figure~\ref{fig:direction} illustrates time delays and sound direction.
	
	%*****************************

	\noindent
	\noindent	
	
	%%%%%%%%%%%%%%%%%%%%%%%%%%%%%%%%%%%%%%%%%%%%%%%%%%%%%%%%%
	%%%%%%%%%%%%%%%%%%%%%%%%%%%%%%%%%%%%%%%%%%%%%%%%%%%%%%%%%%%%
	%%%%%%%%%%%%%%%%%%%%%%%%%%%%%%%%%%%%%%%%%%%%%%%%%%%%%%%%%%%%

	\begin{figure}[t]
		\begin{center}
			\includegraphics[scale=.4]{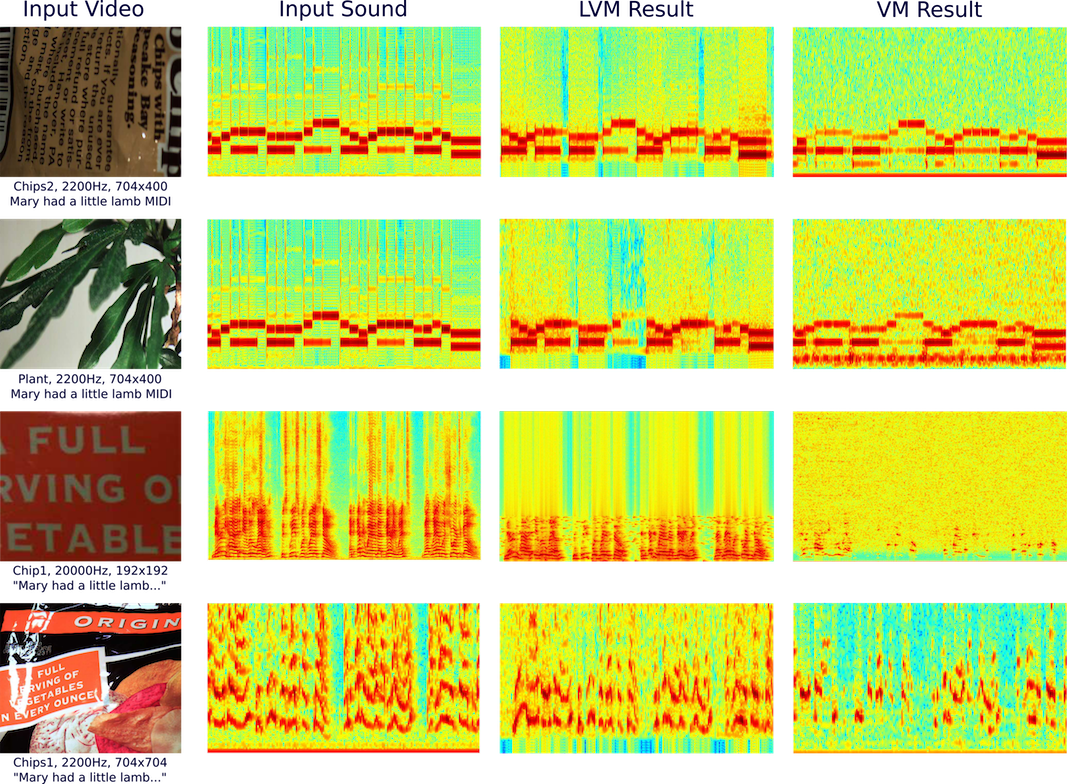}
		\end{center}
		\vspace*{-5mm}
		\caption{Input Video: In the first two videos a melody is played while in the last two videos a human voice is played. Input Sound: Spectrogram of input sound that is played in the video. LVM Result: Spectrogram of our captured sound. VM Result: Spectrogram of sound extracted by~\cite{Davis2014VisualMic}.} 
		\label{fig:soundspect}
	\end{figure}

	\section{Experimental Results}\label{sec:Experimental}
	
	\subsection{Sound Extraction}
	
	Davis \etal~\cite{Davis2014VisualMic} established a benchmark with four high-speed videos. These videos are captured at 2KHz to 22KHz frame-rates. Two videos were captured during a melody playback. Two videos were captured during a speech playback.	All videos are silent and the goal is to recover sound from silent high-speed videos. We compare our technique to Davis \etal~\cite{Davis2014VisualMic} on all four videos. Figure~\ref{fig:soundspect} illustrates the four videos along with the spectrograms of their recovered sound. This figure compares our technique with the previous baseline.
	
	We quantify sound quality using three measures: Perceptual Evaluation of Speech Quality (Table~\ref{table:experimental}), Log-likelihood ratio (Table~\ref{tab:llr}), and Segmental SNR (Table~\ref{tab:ssnr}). We use Segmental SNR and Mean LLR for all four videos. We used Perceptual Evaluation of Speech Quality (PESQ) only for the two human speech videos (PESQ is only designed for speech). We chose these methods because they are invariant to small time-shifts.
	
	Tables \ref{table:experimental}, \ref{tab:llr} and \ref{tab:ssnr}, also compare our technique using both quartic interpolation with quadratic interpolation. Quartic interpolation leads to better sound quality, however, it comes at about twice computational cost. These tables, also compare the quality of sound extraction using only a single pixel (the best pixel). The quality of a single pixel is not as good as the rest of techniques, however, it is noteworthy that a single pixel can produce comprehensible sound.

\begin{table}[H]
		\small
		\begin{center}
			\begin{tabular}{|c|c|c|c|c|c|}
				\hline
				PESQ Score &  \begin{tabular}{@{}c@{}} Global Sound \\ Extraction\end{tabular} & \begin{tabular}{@{}c@{}} Single Pixel \\ Mask \end{tabular} &  VM~\cite{Davis2014VisualMic} &\begin{tabular}{@{}c@{}} Quadratic \\ LVM(ours) \end{tabular} & \begin{tabular}{@{}c@{}} Quartic \\ LVM(ours) \end{tabular}  ~ \tabularnewline
				\hline
				{Chips1, 2200Hz} & 1.168 & 2.385 & 2.486 & 2.708 & \textbf{3.047} \tabularnewline
				\hline
				{Chips1, 20000Hz} & 0.408 & 1.032 &  1.280 &1.878 & \textbf{2.174} \tabularnewline
				\hline
			\end{tabular}
			\vspace*{+2mm}
			\caption{PESQ measurements for recovered sounds. PESQ score ranges between $[0,5]$ and higher scores indicate higher quality.}
			\label{table:experimental}
		\end{center}
	\end{table}

	\begin{table}[H]
		\centering
		\small
		%\makebox[\textwidth][c]{
		%\begin{adjustbox}{max width=\textwidth}
		\begin{tabular}{| c | c | c | c | c | c |}
			\hline
			MeanLLR & \begin{tabular}{@{}c@{}} Global Sound \\ Extraction\end{tabular} & \begin{tabular}{@{}c@{}} Single Pixel \\ Mask \end{tabular} & VM~\cite{Davis2014VisualMic} & \begin{tabular}{@{}c@{}} Quadratic \\ LVM(ours) \end{tabular} & \begin{tabular}{@{}c@{}} Quartic \\ LVM(ours) \end{tabular}   ~
			\\ \hline
			Chips1-2200Hz & 1.394 & 1.453& \textbf{1.019} & 1.027 & 1.021  \\ \hline
			Chips1-20000Hz & 1.272 & 1.250 & 3.566& \textbf{1.233} & 1.277   \\ \hline
			Chips2-2200Hz & 0.847 & 1.242 & 0.591& 0.522 & \textbf{0.407}  \\ \hline
			Plant-2200Hz & 1.027 & 1.190 & 0.853& 0.572 & \textbf{0.510}   \\ \hline
		\end{tabular}
		%\end{adjustbox}
		%}
		\vspace*{+2mm}
		\caption{Log-likelihood ratio measurements for recovered sounds. Lower Log-likelihood ratios indicate higher quality. As shown in the table, quartic and quadratic interpolation can lead to the lowest Log-likelihood ratio. }
		\label{tab:llr}	
	\end{table}
	
	\begin{table}[H]
		\centering
		\small
		%		\makebox[\textwidth][c]{
		\begin{tabular}{| c | c | c | c | c | c |}
			\hline
			SegSNR & \begin{tabular}{@{}c@{}} Global Sound \\ Extraction\end{tabular} & \begin{tabular}{@{}c@{}} Single Pixel \\ Mask \end{tabular} & VM~\cite{Davis2014VisualMic} & \begin{tabular}{@{}c@{}} Quadratic \\ LVM(ours) \end{tabular} & \begin{tabular}{@{}c@{}} Quartic \\ LVM(ours) \end{tabular}~
			\\ \hline
			Chips1-2200Hz & -1.3426  & 0.0764 & 1.2255& 1.2621 & \textbf{1.276} \\ \hline
			Chips1-20000Hz & -2.4603 & -1.1529 & -2.2569& -1.0065 & \textbf{-0.837}  \\ \hline
			Chips2-2200Hz  & -0.0206 & -4.9062& 1.6254 & 1.7942 & \textbf{2.7688}\\ \hline
			Plant-2200Hz & 1.0115 & 0.2378 & 2.7778& 3.6434 & \textbf{4.5339} \\ \hline
		\end{tabular}
		\vspace*{+2mm}
		%		}
		\caption{Comparison of sound extraction techniques using segmental signal to noise ratios (Segmental SNR). a higher segmental SNR scores indicate higher quality. Our technique outperforms the baseline according to Segmental SNR, Log-likelihood ratio and PESQ measures.}
		\label{tab:ssnr}			
	\end{table}
	\subsection{Estimation Of Material Properties}
	
\begin{figure}[t]
		\begin{center}
			\scalebox{.12}{\includegraphics{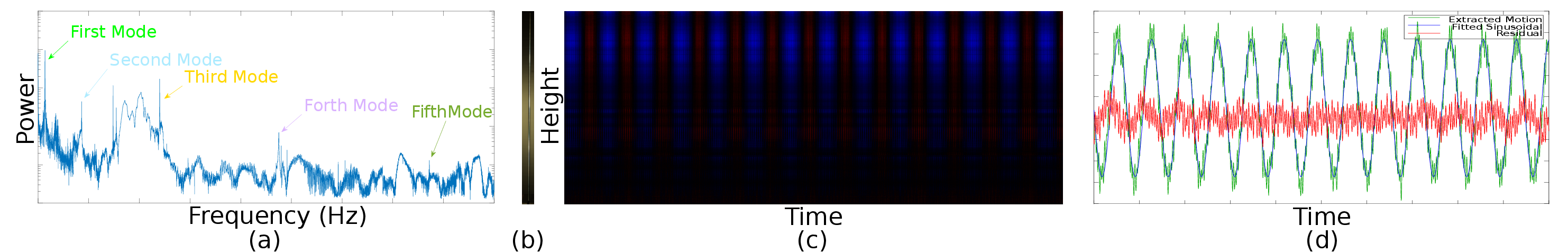}}    
		\end{center}
		\vspace*{-5mm}
		\caption{(a) Vibration of an aluminum rod in frequency domain. Each spike marks a candidate frequency mode. (b) A frame from video of this aluminum rod. (c) Estimated displacement over time at different times. Displacement to the left and right are marked by red and blue. (d) The first mode of vibration and the residual.}
		\label{fig:vibro}
	\end{figure}

	Vibration analysis can help determine certain material properties including Young's modulus. We compare our results on this task with that of~\cite{7299171}. This experiment includes high-speed videos of clamped rods with different materials in two lengths, once clamped to a length of 15 inches and once clamped to a length of 22 inches. There is a connection between a material's Young's modulus and fundamental frequency~(first mode)~\cite{9781468403824}. Figure~\ref{fig:vibro} shows estimated frequency modes of an aluminum rod clamped to a length of 22 inch. Table~\ref{table:vibro} compares our results.

	%*****************************
	
	\begin{table}[h]
		\begin{center}
			\small
			\begin{tabular}{|c|c|c|c|c|}
				\hline
				$\%$ Error & Aluminum & Brass & Copper & Steel \\
				\hline
				$22$ in - Ours & \textbf{-8.23} & \textbf{-0.31} & \textbf{-1.31} & \textbf{-10.01} \\
				$22$ in - Visual Vibrometry & -8.94 & -0.95 & -1.49 & -10.97\\
				\hline
				$15$ in - Ours & \textbf{-14.07} & \textbf{-6.12} & \textbf{-4.21} & -15.17 \\
				$15$ in - Visual Vibrometry & -22.59 & -6.39 & -5.01 & \textbf{-15.09}\\
				\hline
			\end{tabular}
		\end{center}
		\caption{Comparison of methods according to Young's modulus estimation error.}
		\label{table:vibro}
	\end{table}

	%%%%%%%%%%%%%%%%%%%%%%%%%%%%%%%%%%%%%%%%%%%%%%%%%%%%%%%%%%%%
	%%%%%%%%%%%%%%%%%%%%%%%%%%%%%%%%%%%%%%%%%%%%%%%%%%%%%%%%%%%%
	%%%%%%%%%%%%%%%%%%%%%%%%%%%%%%%%%%%%%%%%%%%%%%%%%%%%%%%%%%%%
	\section{Discussion} 
	
	We developed a fast algorithm to extract sound from high-speed videos. We showed that different parts of image vibrate differently and we can improve sound quality by carefully combining local vibrations. Our algorithm has state-of-the-art performance both in terms of sound quality and speed. We also extract further information from videos such as sound direction.
	
	There are a number of possible direction for future work. In theory, if there are multiple sources of sound, it could be possible to separate sounds from different sources. Real-time processing of small motions could have several practical applications in Medical Science, and Engineering. Also our regular memory access pattern opens the possibility to benefit from further hardware accelerators including GPU.
	\section{Acknowledgements}
	
	We greatly thank Dr.~Mostafa Kamali Tabrizi, Dr. S. Ebadolah Mahmoudian and Dr. Mohammad Reza Razvan for their support and their valuable comments. This work was partially supported by National Elites Foundation of Iran.
	
	{

	}
	\end{document}